\documentclass{article} 
\usepackage{iclr2025_conference,times}

\usepackage{amsmath,amsfonts,bm}

\def\eqref#1{equation~\ref{#1}}

\def\1{\bm{1}}

\DeclareMathAlphabet{\mathsfit}{\encodingdefault}{\sfdefault}{m}{sl}
\SetMathAlphabet{\mathsfit}{bold}{\encodingdefault}{\sfdefault}{bx}{n}

\usepackage{hyperref}
\usepackage{url}
\usepackage{booktabs}
\usepackage{multirow}
\usepackage{xcolor}
\usepackage{pdfpages}
\usepackage{orcidlink}
\usepackage{amsfonts,amssymb}
\usepackage{makecell}
\usepackage{fontawesome5}
\usepackage{graphicx}
\usepackage{booktabs}
\usepackage{subcaption}
\usepackage{hyperref}
\usepackage[noabbrev]{cleveref}
\usepackage{float}
\usepackage{wrapfig}
\usepackage{algorithm}
\usepackage{mathrsfs,algorithmic}
\usepackage{fancyhdr}
\usepackage{marvosym}

\title{Less is More: High-value Data Selection for Visual Instruction Tuning}

\author{Zikang Liu{\normalfont\textsuperscript{{1} *}}\thanks{\textsuperscript{*}\:\:\:Equal contribution.}, Kun Zhou{\normalfont\textsuperscript{{2} *}}, Wayne Xin Zhao{\normalfont\textsuperscript{1 \Letter}} \thanks{\textsuperscript{\Letter}\:\:Corresponding author.}, Dawei Gao{\normalfont\textsuperscript{3}}, Yaliang Li{\normalfont\textsuperscript{3}}, Ji-Rong Wen{\normalfont\textsuperscript{1}} \\
\textsuperscript{1}Gaoling School of Artificial Intelligence, Renmin University of China.\\
\textsuperscript{2}School of Information, Renmin University of China. \quad
\textsuperscript{3}Alibaba Group.\\
\texttt{\{jasonlaw8121, batmanfly\}@gmail.com, 
 francis\_kun\_zhou@163.com} \\
 \texttt{\{gaodawei.gdw, yaliang.li\}@alibaba-inc.com, jrwen@ruc.edu.cn} 
}

\iclrfinalcopy 

\makeatletter
\def\thanks#1{\protected@xdef\@thanks{\@thanks
        \protect\footnotetext{#1}}}
\makeatother

\begin{document}

\maketitle

\begin{abstract}
Visual instruction tuning is the key to building large vision language models~(LVLMs), which can greatly improve the task generalization and solving capabilities by learning a mixture of instruction data from diverse visual tasks.
Previous work mostly collects multiple existing visual instruction datasets via heuristic ways for training (even more than a million instructions), which may introduce data redundancy and enlarge the training cost. 
To investigate this issue, we conduct a series of empirical studies, which reveal a significant redundancy within the visual instruction datasets, and show that greatly reducing the amount of instructions from several tasks even do not affect the performance.
Based on the findings, we propose a high-value data selection approach \textbf{TIVE}, to eliminate redundancy within the visual instruction data and reduce the training cost.
In TIVE, we first estimate the instance influence score on its corresponding task, and the task difficulty score, based on the gradient-based influence functions.
Then, we leverage the two kinds of scores to determine the task proportion within the selected visual instruction subset, and select high-value instances for each task, respectively.
Experiments on various LVLMs show that our approach using only about 15\% data can achieve comparable average performance to the full-data fine-tuned model across eight benchmarks, even surpassing it on four of the benchmarks. Our code and data will be publicly released.
\end{abstract}

\section{Introduction}



The advent of large language models~(LLMs)~\citep{Brown2020LanguageMA,Ouyang2022TrainingLM,Touvron2023LLaMAOA,zhao2023survey} has marked significant advancements in the field of artificial intelligence~(AI), exhibiting excellent capabilities in human instruction following, world knowledge utilization, and complex reasoning. 
A surge of recent studies~\citep{Zhu2023MiniGPT4EV,Liu2023VisualIT,Dai2023InstructBLIPTG,liu2023improved} equip LLMs with the vision encoder to empower the capability of processing visual information. 
Through vision-language alignment pre-training and visual instruction tuning, \emph{Large Vision Language Models~(LVLMs)} are created to extend the application of LLMs into multimodal tasks and scenarios.

Visual instruction tuning~\citep{Liu2023VisualIT,Dai2023InstructBLIPTG} is the key technique for improving the task generalization and instruction following capabilities of LVLMs, which relies on a set of visual instructions for fine-tuning.
Therefore, the construction of visual instruction datasets is very crucial for LVLMs.
Typically, there are two widely used ways to construct visual instructions: synthesizing instructions based on LLMs~\citep{Liu2023VisualIT} or transforming existing vision-language datasets into visual instructions~\citep{Dai2023InstructBLIPTG,liu2023improved}. To achieve better performance, existing LVLMs generally combine a mixture of visual instructions from different domains or tasks, to compose a large-scale visual instruction dataset.
The LVLMs fine-tuned on these mixtures of visual instructions have shown remarkable performance on massive downstream multimodal benchmarks.

\begin{wrapfigure}{R}{0.5\textwidth}
     \centering
\includegraphics[width=0.5\textwidth]{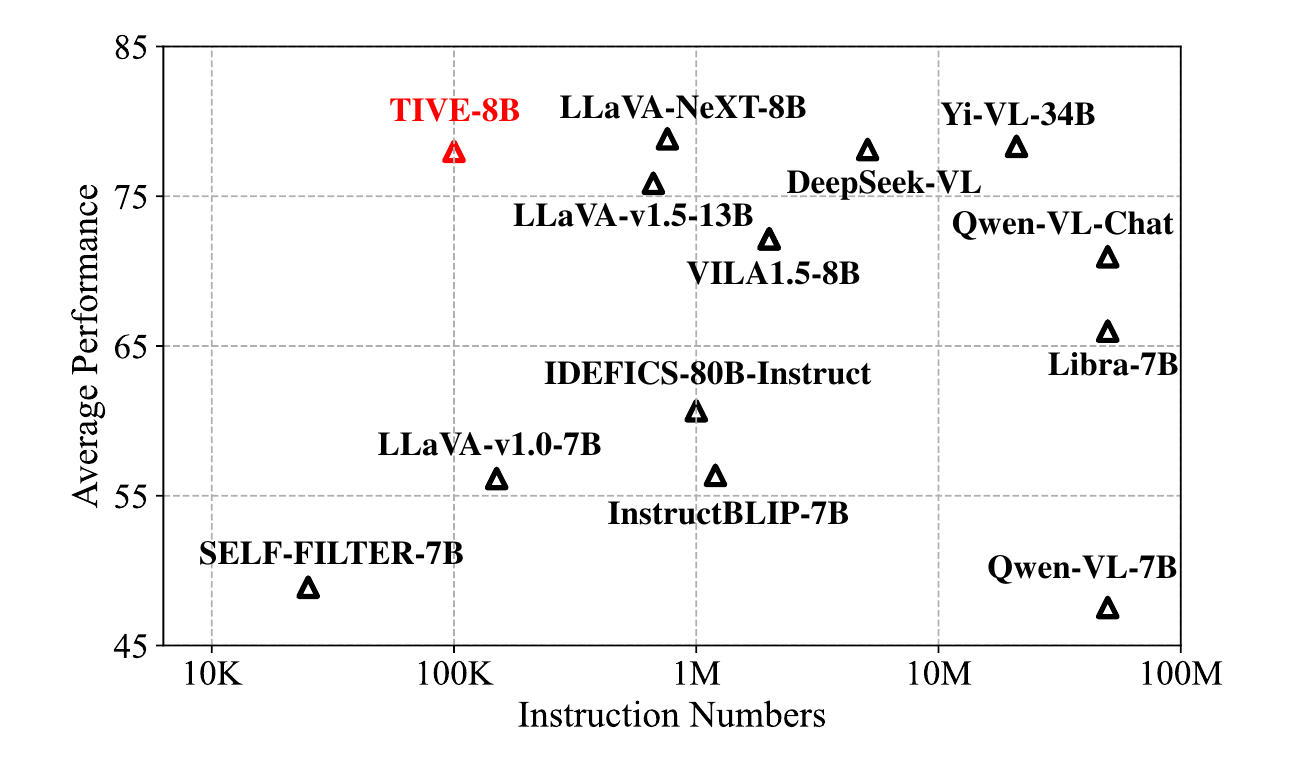}
\caption{A comparison of TIVE-8B with other open-source models in terms of the instruction data scale and average benchmark performance on MME, SEED-Bench, MMBench, ScienceQA.}
\label{fig:intro}
\end{wrapfigure}

However, such a mixture of instructions may also introduce significant data redundancy, leading to increased training costs and potentially overfitting risk.
To investigate the redundancy issue, we first conduct an empirical study on the visual instruction dataset of state-of-the-art open-source LVLM, \ie, LLaVA-1.5~\citep{liu2023improved}, by reducing the instruction amount of a certain task and then evaluating the performance.
The results show that the reduction of instruction data only leads to slight or even no performance decline across most benchmarks, indicating that there exists redundancy within the used visual instructions.
Therefore, it is promising to mitigate this redundancy by selecting a small set of representative data samples. 
Furthermore, we also find that the degree of redundancy varies across different tasks. It suggests that the contribution of each task should be considered when performing the redundancy elimination.

To this end, in this paper, we propose a data selection approach for visual instruction tuning, namely \textbf{TIVE}, based on \emph{Task and Instance Value Estimation}. 
The key motivation is to estimate the value of each instance and then select the high-value ones, based on its influence on LVLM fine-tuning process.
According to the influence function theory~\citep{Pruthi2020EstimatingTD}, the influence of an instance on the training process can be estimated by its gradient similarities with other instances.
However, due to the large-scale parameters of LVLMs, the computation of gradient similarity may cause unaffordable cost.
Besides, since the goal of visual instruction tuning is to learn the solving capability for diverse tasks, it is necessary to measure the influence of task learning~\citep{Pruthi2020EstimatingTD,xia2024less}, instead of only cross-instance influence.

In TIVE, we adjust the gradient computation and influence estimation strategies, to better adapt into visual instruction tuning of LVLMs.
To reduce the cost, we only leverage the gradients of the LoRA~\citep{Hu2021LoRALA} matrices from LLM for influence estimation. These parameters are the key components for learning visual understanding and instruction following capabilities, hence their gradients would be informative features.
To focus on task learning, we estimate the contribution of each instance to its corresponding task, by computing the average influence of each instance on all other in-task instances, namely \emph{instance influence score} to help distinguish the most useful instances.
Then, we measure the difficulty of each task for LVLM to learn, by computing the average self-influence of all its contained data instances, as the \emph{task difficulty score} to help determine the task data proportion.
Guided by the above scores, we can select the high-value instances to fine-tune the LVLM, for efficiently and effectively learning all the involved tasks within the visual instruction dataset.

To demonstrate the effectiveness of our approach, we apply our data selection method into several SOTA LVLMs and widely-used instruction datasets, and perform evaluation on eight benchmarks.
By only using the selected 15\% subset from the visual instruction dataset, the fine-tuned LVLMs can achieve comparable performance to the full-data fine-tuned model, even outperforms it on four benchmarks.
As shown in Figure~\ref{fig:intro}, our TIVE-8B (based on LLaVA-LLaMA3-8B) can reach the SOTA performance with much fewer instructions than SOTA methods.

\section{Redundancy Analysis on Visual Instruction Data}\label{empirical}

In this section, we conduct an empirical study to examine: (1) whether data redundancy exists in existing visual instruction datasets, and (2) whether the degree of redundancy differs in different task instructions. 


\begin{figure}
\centering
\begin{subfigure}{.32\textwidth}
    \centering
    \includegraphics[width=.98\linewidth]{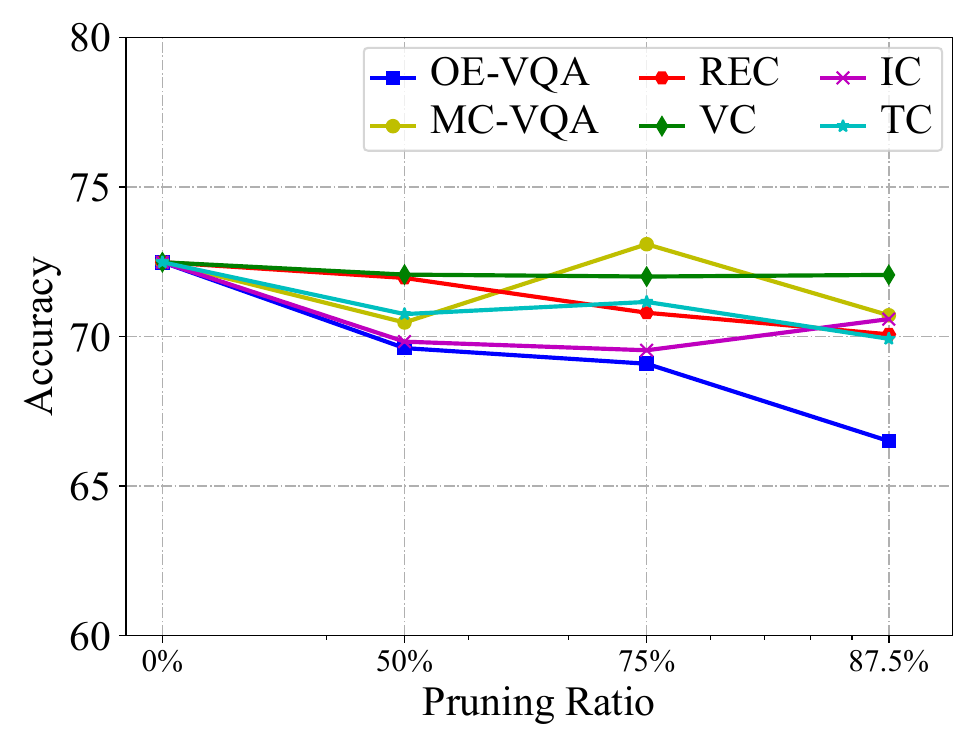}  
    \caption{MME-P}
\end{subfigure}
\begin{subfigure}{.32\textwidth}
    \centering
    \includegraphics[width=.98\linewidth]{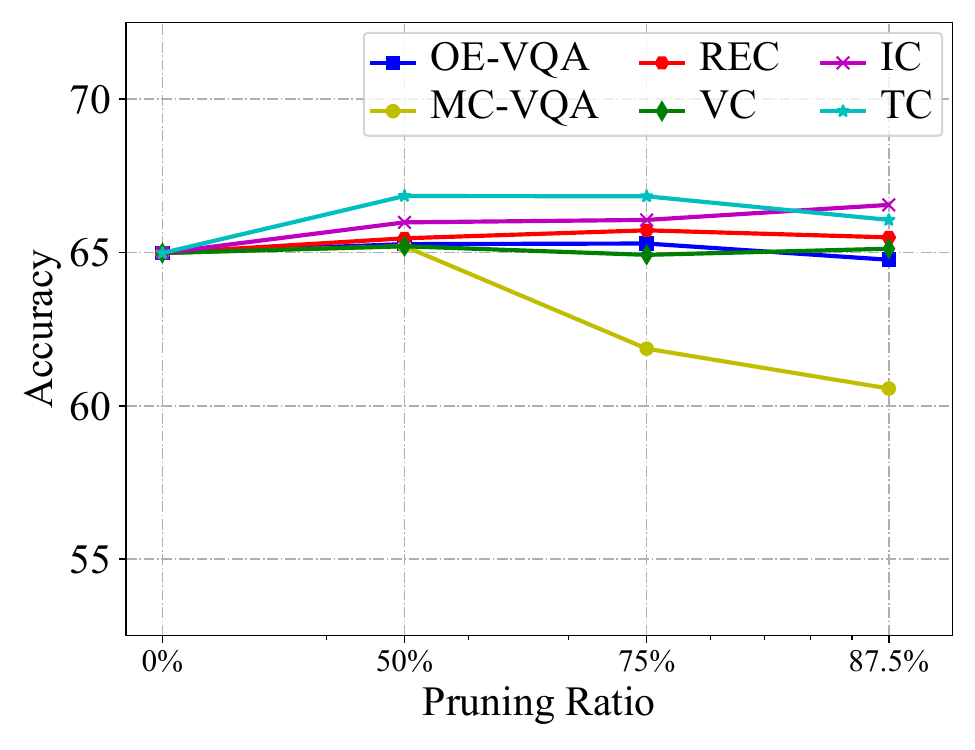}  
    \caption{MMBench}
\end{subfigure}
\begin{subfigure}{.32\textwidth}
    \centering
    \includegraphics[width=.98\linewidth]{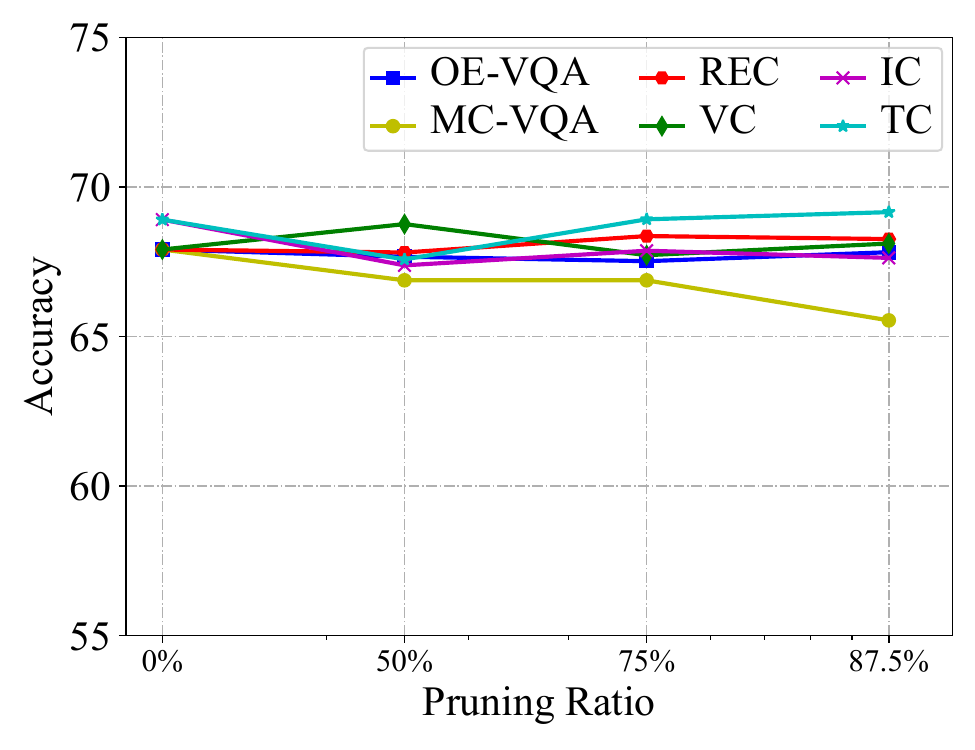}  
    \caption{ScienceQA~(Image)}
\end{subfigure}
\caption{Evaluation results after pruning the amount of visual instructions from one task. Pruning 87.5\% data for most tasks only leads to slight performance changes on three benchmarks.}
\label{figure-empirical}
\end{figure}

\subsection{Analysis Setup}
Given a mixture of visual instruction datasets for training LVLMs, we prune the amount of visual instructions from a certain task and then examine the performance change after fine-tuning with the adjusted instruction dataset. 
In this experiment, we mainly study the used instruction dataset for training the SOTA open-source LVLM, LLaVA-1.5~\citep{liu2023improved}.

\paragraph{Backbone Model.}
We choose the LLaVA-1.5~\citep{liu2023improved} model after cross-modal alignment training as the backbone model (without instruction-tuning), which has been trained on more than 500k image-text pairs.
It incorporates CLIP~\citep{radford2021learning} as the visual encoder and Vicuna-v1.5~\citep{vicuna2023} as the LLM, and further leverages two linear layers for mapping the encoded visual features to the latent space of LLM.

\paragraph{Visual Instruction Dataset.}
LLaVA-1.5 has been fine-tuned on a mixture of instruction datasets from different tasks. To ensure internal consistency across different tasks, only one dataset for each type of task will be selected. We select VQAv2~\citep{goyal2017making} dataset for 
Open-Ended Visual Question Answering~(OE-VQA), A-OKVQA dataset~\citep{schwenk2022okvqa} for Multi-Choice Visual Question Answering~(MC-VQA), RefCOCO~\citep{mao2016generation,kazemzadeh2014referitgame} dataset for Referring Expression Comprehension~(REC), LLaVA-1.0~\citep{Liu2023VisualIT} dataset for Visual Conversation~(VC), CC3M~\citep{sharma2018conceptual} dataset for Image Caption~(IC), and ShareGPT~\citep{zheng2023judging} dataset for Textual Conversation~(TC). \autoref{apdx-data} contains details about the datasets.

To investigate the redundancy issue in visual instruction datasets, we gradually halve the number of instructions from each task, then fine-tune the backbone model on the new instruction set and finally compare the performance change. 
For all experiments, we follow the default experimental configuration of LLaVA-1.5. 

\paragraph{Evaluation Benchmark.}
To conduct a comprehensive empirical analysis, we evaluate the fine-tuned LVLMs on the three commonly-used benchmarks: MME-P~\citep{fu2023mme}, ScienceQA~\citep{lu2022learn}, and MMBench~\citep{liu2023mmbench}. Detailed descriptions
of these benchmarks are available in \autoref{apdx-evaluation}.

\subsection{Results and Findings}
According to the results in \autoref{figure-empirical}, we list the main findings as follows:

First, \emph{there exists a significant redundancy in visual instruction datasets}. We can observe that decreasing the amount of instruction data only leads to slight performance drop in most cases. 
For example, reducing the number of VC would not significantly affect the model's performance across all benchmarks, and even lead to improvement on ScienceQA using 50\% of data.
It indicates that not all the used instruction datasets are indispensable.

Second,  \emph{for each task, the redundancy degree of different instruction datasets differs}.
For OE-VQA and MC-VQA, reducing their instruction number leads to relatively significant performance degradation, \eg 8\% on MME-P and 7\% on MMBench using a pruning ratio of 87.5\%, respectively.
While pruning task instructions from VC leads to minimal decline on most of the benchmarks. It indicates that different task instructions contribute to the model's final performance differently. Therefore, it is necessary to estimate the value of each task, for helping set a more proper pruning ratio and mixing proportion for all the tasks.




\section{Approach}
\begin{figure}[t]
    \centering
    \includegraphics[width=0.9\textwidth]{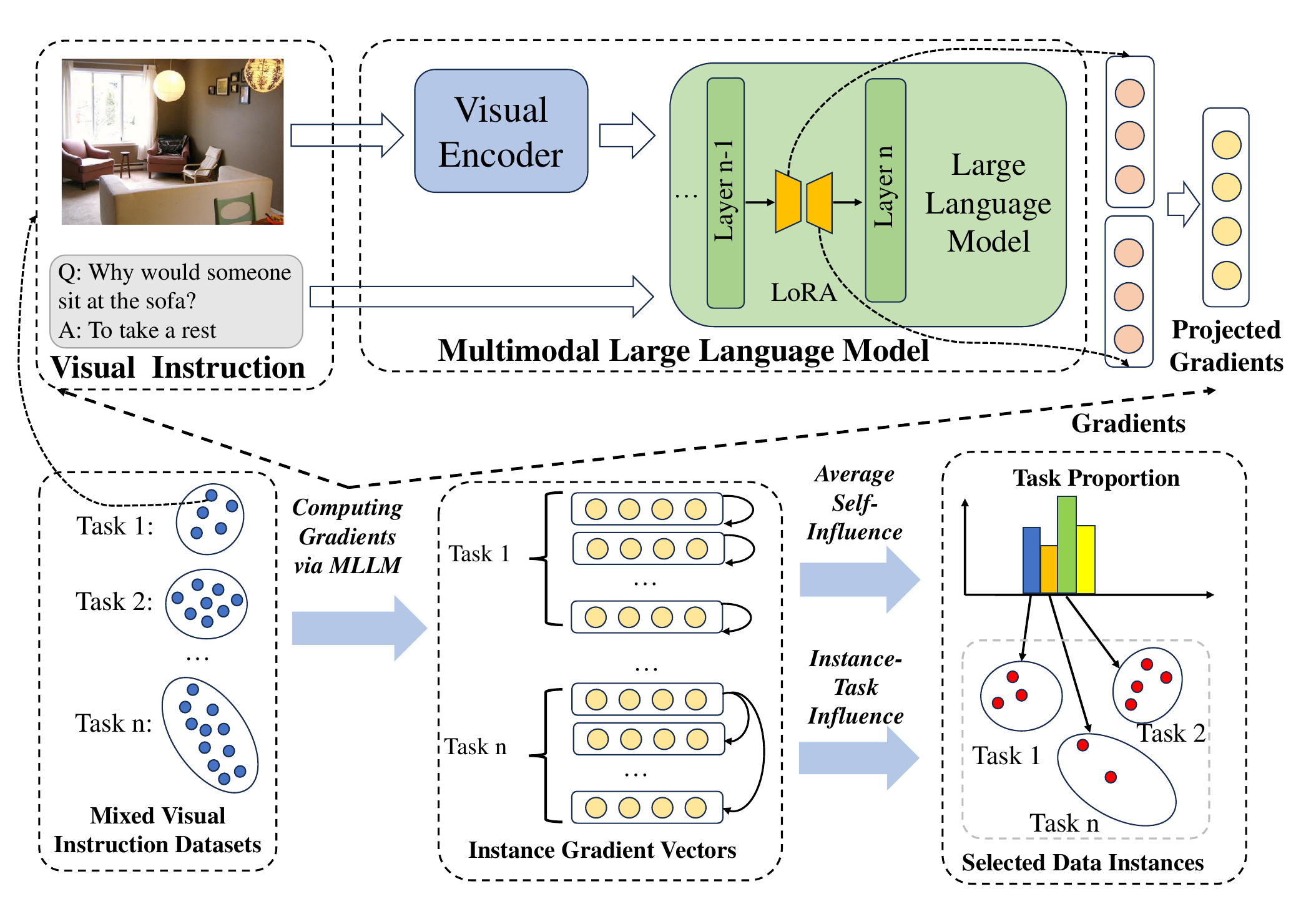}
    \caption{The illustration of our proposed approach. We utilize the gradient vectors from the LoRA parameters of the LLM, to compute the task difficulty and instance influence scores. Then, these scores are leveraged to determine the task data proportion and instance selection probability.}
    \label{fig:main}
\end{figure}

In this section, we present our approach \textbf{TIVE}, to reduce the redundancy of visual instruction data.
Based on the findings in~\autoref{empirical}, it is necessary to consider the contribution degree to the learning of diverse tasks during fine-tuning LVLMs.
Specially, we consider measuring both task difficulty and instance influence scores for helping select visual instruction data. 
Based on the two kinds of estimated scores, we design the data selection process, to sample a small high-value visual instruction subset for efficiently and effectively fine-tuning LVLMs. 
We show the details of TIVE in \autoref{fig:main}.


\subsection{Problem Formulation}
The elimination of dataset redundancy aims to select a high-quality subset from a large dataset suffering the redundancy issue. The selected subset should contain relatively few but informative samples, to ensure the performance of the models trained on it.
In this work, we focus on reducing the redundancy of the visual instruction data pool $\mathcal{D} = \left\{ D_1,...,D_n \right\}\ $, which is a mixture of multiple highly diverse instruction datasets from different tasks. Each dataset comprises a set of instruction samples, denoted as $D_i = 	\left\{ s_1,...,s_n \right\}\ $ .
Our goal is to select a data subset $\mathcal{D}_T$ from the visual instruction data pool for fine-tuning LVLMs. We use $|\mathcal{D}_T|$ to denote the target size of the selected subset. 

Specially, we select the data subset from two perspectives, with the help of a pre-learned reference model trained on the sampled small set of the visual instruction data.
First, we estimate the value of each task and rely on its difficulty to determine their proportions within the final subset $\mathcal{D}_T$.
Second, we estimate the value of each instance within each task $D_i$ to select the most useful instances for this task.

\subsection{Estimating Task Difficulty and Instance Influence}
In this part, we present how we measure the task difficulty and instance influence scores based on the influence on fine-tuning LVLMs.
According to the influence formulation~\citep{Pruthi2020EstimatingTD}, the influence of a training instance $s$ on the another instance $s'$ can be denoted as:
\begin{align}
  \texttt{Inf}(s, s') = \nabla l(s, \theta) \cdot \nabla l(s', \theta),
\end{align}
where $\theta$ and $\nabla l(s, \theta)$ denote the parameters of the LVLM and their gradients, respectively.
Based on it, we devise two formulations for estimating the influence of each instance on learning its corresponding task, and measuring the difficulty of learning each task, respectively.

\paragraph{Instance Influence Estimation.}
To efficiently learn each task during visual instruction tuning, we aim to obtain the contribution of each task instance, to help select a small proportion of training samples which are highly important for the task learning. 
Our motivation is that if an instance has a higher positive influence on the learning of all other instances within the task, it can be regarded as a higher-value instance for helping learn the task and should be selected. 
Therefore, given an instance $s$ from task set $D_i$. we compute the average influence of the instance on all other instances from its affiliated task, denoted as:
\begin{align}
v^i_{s} = \frac{1}{|D_i|}\sum_{s' \in D_i \setminus s} \frac{\nabla l(s, \theta) \cdot \nabla l(s', \theta)}{|\nabla l(s, \theta)| |\nabla l(s', \theta)|}.
\end{align}


We normalize the gradients to mitigate the impact caused by abnormally large gradient values.
By this way, we can compare the influence of different instances within each task, and select the high-value ones for training.

\paragraph{Task Difficulty Estimation}\label{sec-taskvalue}
According to our findings in \cref{empirical}, the impact of pruning different task instruction amount also differs in the LVLM performance. 
It is because not all the involved tasks are so hard that require such number of training instances, and it is promising to prune their data amount for reducing redundancy.
Therefore, we aim to measure the difficulty of all the tasks within the visual instruction dataset, to  adjust their proportion in the selected subset. 
Concretely, we employ the average self-influence score of all the in-task instance, to measure the task difficulty.
Self-influence is to estimate the influence of training an instance on learning itself, denoted as $\nabla l(s, \theta) \cdot \nabla l(s, \theta)$.
A higher self-influence score indicates that the instance is hard to learn~\citep{bejan2023make}, as it leads to large gradient values.
By averaging the self-influence scores of all instances from each task, we can estimate the overall difficulty of a task as:
\begin{align}
  v_i^{t} = \frac{1}{|D_i|}\sum_{s \in D_i}\nabla l(s, \theta) \cdot \nabla l(s, \theta).
  \label{eq:task}
\end{align}
Based on the task difficulty score, we can determine the proportion of all the task data within the selected visual instruction subset.
In this way, the difficult task should be assigned with a larger proportion of selected data, while the redundant data within the easy tasks should be removed more.

\subsection{Data Subset Selection}
In this section, we introduce how we obtain the gradient features and select a small data subset based on the proposed data value measurements.  

\paragraph{Gradient Features Computation.}
Firstly, to efficiently compute the gradient features, we train a reference model with LoRA~\citep{Hu2021LoRALA} using a small amount of instruction data. In this way, the reference model can be warm-up to learn the visual instruction following capability, and has not overfitted to the distribution of the whole visual instruction dataset.
Thus, the gradients from the reference model can store useful information about visual instruction tuning for following influence estimation.
After training the reference model, we can obtain the gradient features through backward propagation. To save storage and computation, we follow existing work~\citep{Pruthi2020EstimatingTD} to reduce feature dimensions with random projection. Such projection often preserves the inner products~\citep{johnson1984extensions}, ensuring the effectiveness of the projected gradient features.

\paragraph{Selecting Data based on Estimated Values.}
After obtaining the task-level and instance-level data values, we can select the subset from the visual instruction data pool. 
First, we use the task-level value to determine the proportion for each task in the data subset. 
The target data subset $\mathcal{D}_T = \{ D_1^{'},...,D_n^{'} \}\ $  contains the same number of task datasets as the original data pool, but changes the total amount and task proportion.
For each task subset $D_i^{'}$, we compute its data proportion within the target data subset as $p_i^{'} = \frac{v^t_i}{\sum_{j=1}^{n}v^t_j}$.
\vspace{-0.5mm}
where $v^t_i$ is the estimated task-level value. Then, we rely on the instance-level value to sample $|D_i^{'}|$ instances from the original visual instruction dataset.
Here, we directly employ the softmax function to map the instance-level value to a sampling weight distribution. We use a hyperparameter $\lambda$ to control the temperature of the weight distribution. For all the tasks, we sample the instances based on the above weight distribution, and merge all the datasets to compose our final selected data subset.


\section{Experiments}
\begin{table*}[t]
\small
\centering
\caption{A comparison between TIVE and other baseline approaches for data selection on several downstream benchmarks. Benchmark names are abbreviated due to space limits. MME-P: MME-Perception, MME-C: MME-Cognition,  SEED-I: SEED-Bench~(Image), MMB: MMBench, MMB-CN: MMBench~(Chinese), SQA: ScienceQA, SQA-I: ScienceQA~(Image). * indicates our reimplemented results. Rel. represents the average relative performance compared to baseline model. Improvement over best represents the relative improvement of TIVE over the best performance among other baseline approaches. \textbf{Bold} and \underline{underline} fonts indicate the best and second best performance on the task.}
\resizebox{\textwidth}{!}{
\begin{tabular}{@{}lr|ccccccccc@{}}
\toprule
Method                                 & 
\# Ins  & MME-P & MME-C & SEED-I & MMB & MMB-CN & SQA  & SQA-I & POPE & Rel.\\ \midrule
BLIP-2          &  -    & 1293.8   & -      & -  & -           &  -    & -          & 61.0  & 85.3    & -                                                  \\
InstructBLIP-7B &  1.2M  & -     & -         & -  &  36.0         & 23.7             &  -   & 60.5     & -           & -                               \\
Shikra                          & 5.5M    & -      & -      & -  & 58.8            & -          & - & -         & -                 & -                         \\
IDEFICS-80B         & 1M  & -      & -        & -  &  54.5          & 38.1             &  -   & -           & -  & -                                       \\
Qwen-VL        & 50M  & - & -  & -            & 38.2 &  7.4          & -             &  67.1   & -    & -
\\
Qwen-VL-Chat         & 50M  & 1487.5  & -  & -            & 60.6 &  56.7          & -             &  68.2   & -    & -
\\
InstructionGPT-4    &  0.2K  &  463.3  & -  &   -   & 31.4   &   -   &  -   &  -   &  -  & - \\ 

SELF-FILTER     &  25K   &   955.6   & -  &  47.5   &   38.5   &   -   &   59.4   &   -   &   -  & -
\\ 
\midrule
\textbf{\small{Backbone model}} \\
LLaVA-1.5                & 665K & \textbf{1510.7}      &  \underline{311.9*}     & \textbf{66.1}  &  \underline{64.3}               & \textbf{58.3}             &     69.4*     & 66.8    & \textbf{85.9}   & 100.0\%                                    \\ \midrule
\textbf{\small{Our experiment}} \\
Random                                     & 100K     & 1386.5  & 271.3  & 61.9           & 61.8 & 54.5            & 69.8        & 68.4      & 83.9   & 95.2\%                        \\

Length                                  & 100K     & 1413.0  & 266.1  & 61.2           & 59.3 & 53.9            & 71.1        & \underline{69.2}      & 83.3      & 94.8\%                        \\
Perplexity                                   & 100K     & 1393.3  & 260.7  & 61.3          & 62.3 & 55.0             & 70.5        & 67.9         & 83.6       & 94.9\%                     \\
GraNd                                 & 100K     & 1400.5  & 287.1  & 62.3           & 62.9 & 54.3             & \underline{71.4}        & 68.4           & 82.5    & 96.3\%                       \\
EL2N                                   & 100K     & 1356.5  & 294.7  & 61.9           & 61.6 & 56.1             & 70.2        & 66.2         & 84.6    & 95.5\%                         \\
TIVE~(ours)                                 & 100K   &  \underline{1433.0} & \textbf{322.1}   & \underline{63.2} & \textbf{65.0}           & \underline{58.2}   & \textbf{72.2}             & \textbf{70.6}        & \underline{85.6} & 100.3\%                                   \\

Improve over best     & -   &   1.4\%  & 9.3\%  &  1.4\%  &  3.3\%  &  3.7\%  &  1.1\%   & 
 2.0\%   &  1.2\%  & 4.0\% \\

\bottomrule
\end{tabular}}
\label{tab:main-table}
\end{table*}

\subsection{Experiment Setup}


We conduct extensive experiments on TIVE across various models and datasets. The models include LLaVA-1.5-7B, LLaVA-1.5-13B, LLaVA-Phi-3-4B, and LLaVA-LLaMA3-8B. The datasets include LLaVA-1.5 instructions, SVIT-Mix~\citep{zhao2023svit} instructions and, Mini-Gemini~\citep{li2024mini} instructions. Since LLaVA-1.5-7B and LLaVA-1.5-13B takes Vicuna-7B and Vicuna-13B as their LLM backbone, we denote these two models as LLaVA-Vicuna-7B and LLaVA-Vicuna-13B in some experiments. More information about the training datasets, evaluation benchmarks, and implementation details are presented in \autoref{apdx-data}, \autoref{apdx-evaluation}, and \autoref{apdx-implement}, respectively.

\subsection{Baselines}
We compare our methods with several baselines for data selection: (1) \emph{Random Selection} selects data randomly; (2) \emph{Instruction Length} utilizes length of instruction to determine the importance of an instruction sample; (3) \emph{Perplexity} computes the perplexity score of an instruction sample to measure its importance; (4) \emph{GraNd}~\citep{paul2021deep} measures the importance of each sample by the L2-norm of the gradient caused by each sample; (5) \emph{EL2N}~\citep{paul2021deep} measures the importance of each sample by the L2-norm of the error vector of each sample. 
The EL2N scores are primarily used for estimating sample importance in image classification tasks. To adapt it for visual instruction tuning, we compute the error vector for each token in each sample, and then compute the final EL2N score by averaging norms of all error vectors.

\subsection{Main Results}

We present the comparison of TIVE with other baseline methods on LLaVA-1.5 in~\autoref{tab:main-table}, the results of TIVE across different LVLMs in ~\autoref{tab:ablation-3}, and the results of TIVE across different instruction datasets in ~\autoref{tab:ablation-4}.
We present analyses of the results as follows:

\paragraph{Comparison of TIVE with other baseline methods.}
In ~\autoref{tab:main-table}, we compare TIVE with several baseline methods on 8 benchmarks. First, we observe that the traditional data selection approaches~(GraNd and EL2N) perform slightly better than random selection. A possible reason is that these approaches indeed select valuable data, but are also more vulnerable to the data noise, resulting in a limited improvement. For the data selection approaches used in LLM instruction tuning~(Length and Perplexity), the performances across several benchmarks are even worse than random selection. We discover that these approaches mostly focus on selecting samples which have a high influence on improving the model's generation ability, which leads to minor enhancement on the model's ability on visual understanding. It is clear that our approach significantly outperforms all other baselines and achieves consistently promising results across all benchmarks under a limited data setting. With only \emph{15\% of the instruction data}, our approach can achieve \emph{100.3\%} average performance on all benchmarks compared to the LLaVA-1.5 model, even surpass the performance of LLaVA-1.5 in four benchmarks. These results show that our proposed approach can effectively address the issues of data redundancy within LLaVA-1.5 instructions.

\begin{table*}[t]
  \caption{The performance of TIVE across different LVLMs. \# Samp indicates the sampling ratio.
  }
  \label{tab:ablation-3}
  \centering
  \small
  \scalebox{0.8}{
  \begin{tabular}{>{\centering\arraybackslash}p{.11\linewidth}|>{\centering\arraybackslash}p{.09\linewidth}|>{\centering\arraybackslash}p{.07\linewidth}|*{4}{>{\centering\arraybackslash}p{.08\linewidth}}>{\centering\arraybackslash}p{.07\linewidth}>{\centering\arraybackslash}p{.08\linewidth}>{\centering\arraybackslash}p{.07\linewidth}>{\centering\arraybackslash}p{.07\linewidth}}
    \toprule
    Model & Method   &  \# Samp     & MME-P  &  MME-C & SEED-I & MMB & SQA  &  SQA-I  &  POPE  & Rel.\\
    \midrule
        \multirow{4}*{\makecell[c]{LLaVA-\\ Vicuna-7B}} &   - &  100\%  &   \textbf{1510.7}   &   \underline{311.9}    &   \textbf{66.1}  &  64.3  &  69.4  &  66.8  &  \textbf{85.9}  & 100.0\% \\
    ~ & Random  &  15\%   &  1386.5  &   271.3    &   61.9  &  61.8  &  69.8  &  68.4  &  83.9  &  95.2\%  \\
    ~ & TIVE  &  15\%   &  1433.0  &   \textbf{322.1}    &   63.2  &  \underline{65.0}  &  \textbf{72.0}  &  \textbf{70.6}  &  \underline{85.6}  &  100.3\%  \\
     ~ & TIVE    &   30\%   &   \underline{1467.2}   &   309.8    &   \underline{64.4}  &  \textbf{66.5}  &  \underline{71.4}  &  \underline{70.1}  &  85.2
 &  100.6\%    \\
     \midrule
    \multirow{4}*{\makecell[c]{LLaVA-\\ Vicuna-13B}} &   - &  100\%  &   \underline{1531.3}   &   295.4    &   \textbf{68.2}  &  \underline{67.7}  &  \underline{74.4}  &  71.6  &  85.9  & 100.0\% \\
    ~ & Random  &  15\%   &  1456.6  &   \underline{307.1}    &   63.4  &  64.9  &  73.5  &  69.4  &  85.5  &  96.6\%  \\
    ~ & TIVE  &  15\%   &  1502.9  &   \textbf{336.1}    &   65.3  &  66.1  &  \textbf{74.5}  &  \textbf{72.2}  &  \underline{86.3}  &  100.5\%  \\
     ~ & TIVE    &   30\%   &   \textbf{1545.4}   &   298.6    &   \underline{65.6}  &  \textbf{68.8}  &  74.2  &  \textbf{72.2}  &  \textbf{86.5}
 &  100.1\%    \\
     \midrule
     \multirow{4}*{\makecell[c]{LLaVA-\\ Phi-3-4B}} &   - &  100\%  &   \textbf{1440.8}  &   301.6    &  \textbf{66.7}  &  \underline{67.9 } &  81.0  &  \underline{73.6}   &  \textbf{85.1}  & 100.0\%  \\
    ~ & Random  &  15\%   &  1329.1  &   295.4    &   63.1  &  64.0  &  80.2  &  71.2  &  82.8  &  95.7\%  \\
    ~ & TIVE  &  15\%   &  1386.9  &   \underline{306.4}    &   63.9  &  66.0  &  \underline{81.2}  &  73.5  &  \underline{84.1}  &  98.0\%  \\
     ~ & TIVE    &   30\%   &   \underline{1425.0}   &   \textbf{338.2}    &   \underline{65.1}  &  \textbf{68.5}  &  \textbf{81.8}  &  \textbf{74.3}  &  83.8 
 &  100.9\%    \\
     \midrule
     \multirow{4}*{\makecell[c]{LLaVA-\\ LLaMA3-8B}} &   - &  100\%  &   \textbf{1569.4}   &   \textbf{338.6}    &   \textbf{68.8}  &  \underline{71.2}  &  77.2  &  73.5  &  \textbf{85.7}  &  100.0\% \\
    ~ & Random  &  15\%   & 1495.8  &   318.2    &   65.2  &  67.9  &  80.4  &  \underline{75.4}  &  83.3  &  97.4\%  \\
    ~ & TIVE  &  15\%   &  1511.4  &   \underline{331.1}    & 67.4  &  69.8  &  \textbf{81.6}  &  \textbf{75.7}  &  \underline{84.9}  &  99.5\% \\
     ~ & TIVE    &   30\%   &   \underline{1560.3}   &   322.9    &   \underline{68.1}  &  \textbf{72.0} &  \underline{80.5}  &  74.1    & 84.6 
 &  100.2\%     \\
    
  \bottomrule
  \end{tabular}}
\end{table*}

\paragraph{Performance of TIVE across different LVLMs.} \autoref{tab:ablation-3} shows the performance of TIVE on different LVLMs. We find that under the same sampling ratio~(15\%), our approach significantly outperforms the random baseline across all LVLMs on all benchmarks, achieving an average improvement of at least 2.3\%. Simultaneously, when the sampling ratio is increased to 30\%, our approach achieves better average performance than full data performance across all models, proving that TIVE successfully eliminates redundancy in visual instruction data and is effective across different LVLMs. Furthermore, we discover that under a low sampling ratio~(15\%), LLaVA-Vicuna-13B achieves the best average relative performance~(100.5\%), while LLaVA-Phi-3-4B achieves the worst~(98.0\%). This indicates that LVLMs with a larger LLM backbone have a relatively better average performance under less data, which is consistent with the results for LLM on language instruction tuning scenarios.

\paragraph{Performance of TIVE across different instruction datasets.} We present the results of TIVE on two other instruction datasets in \autoref{tab:ablation-4}. We observe that TIVE remains effective on different instruction datasets. On the SVIT-Mix dataset, it significantly outperforms other baselines in five out of six benchmarks, and surpasses the full data performance in three out of the six benchmarks. On the Mini-Gemini dataset, TIVE shows more advantage over the other baseline methods, and the average performance of TIVE on these benchmarks is better than the full data performance. Considering that the Mini-Gemini dataset has a larger number of instructions, TIVE may be more effective at eliminating redundancy when dealing with a substantial amount of instructions. These results demonstrate the effectiveness of TIVE across different instruction datasets. 



\begin{table*}[t]
  \caption{The performance of TIVE across different instruction datasets.
  }
  \label{tab:ablation-4}
  \centering
  \scalebox{0.8}{
  \begin{tabular}{>{\centering\arraybackslash}p{.15\linewidth}>{\centering\arraybackslash}p{.08\linewidth}|*{6}{>{\centering\arraybackslash}p{.1\linewidth}}>{\centering\arraybackslash}p{.08\linewidth}}
    \toprule
    Method   &  \# Samp   &  MME-P  &  MME-C  & MMB 
  &  SQA & SQA-I  &  POPE  &  Rel.\\
  \midrule
  \multicolumn{9}{c}{\textbf{LLaVA-1.5}} \\
  \midrule
    Baseline &  100\%  &   \textbf{1510.7}   &   \underline{311.9}     &  \underline{64.3}  &  69.4  &  66.8  &  \textbf{85.9}  & 100.0\% \\
    \midrule
    TIVE  &  15\%   &  \underline{1433.0}  &   \textbf{322.1}     &  \textbf{65.0}  &  \textbf{72.0}  &  \textbf{70.6}  &  \underline{85.6}  &  100.3\%  \\
    Random  &  15\%   &  1386.5  &   271.3     &  61.8  &  69.8  &  68.4  &  83.9  &  95.2\%  \\
    Length  &  15\%  &  1413.0  &  266.1  &  59.3  &  \underline{71.1}  &  \underline{69.2}  &  83.3  &   94.8\%  \\

  \midrule
  \multicolumn{9}{c}{\textbf{SVIT-Mix}} \\
  \midrule
    Baseline    &   100\% &   \textbf{1443.5}      &   \underline{306.1}
   &  \textbf{67.3}   &  \underline{70.2}  &  \underline{68.0}  &  \textbf{85.3} & 100.0\%    \\
    \midrule
      TIVE &  15\%  &   1391.7   &   \textbf{306.8}    &  \underline{65.8}  &  \textbf{72.3}  &  \textbf{71.2}  &  \underline{84.3}  &  99.8\% \\
      Random   &  15\%  &  \underline{1402.9}  &  288.8  &  60.2  &  69.6  &  65.7  &  83.8  &  96.0\% \\
      Length   &   15\%  &  1366.5  &  301.1  &  61.3  &  \underline{70.2}  &  67.1  &  84.2   &  96.8\%  \\
      \midrule
  \multicolumn{9}{c}{\textbf{Mini-Gemini}} \\
  \midrule
    Baseline    &   100\% &   \textbf{1538.4}      &   \underline{324.9}
   &  \textbf{68.1}    &  \underline{72.0}  &  \underline{69.9}  &  \underline{85.1}   & 100.0\%  \\
    \midrule
      TIVE &  15\%  &   \underline{1506.9}   &   \textbf{345.4}    &  \underline{67.9}  &  \textbf{72.6}  &  \textbf{71.1} &  \textbf{85.4}  & 101.2\%\\
      Random   &  15\%  &  1404.8  &  305.4  &  62.2  &  71.1  &  69.2  &  84.9  &  95.7\% \\
      Length   &   15\%  &  1403.3  &  313.2  &  62.1  &  70.3  &  67.9  &  83.7  &  95.4\%   \\
    
  \bottomrule
  \end{tabular}}
\end{table*}


\subsection{More Detailed Analysis}

\paragraph{Effectiveness of Data Value Measurements.} We conduct a series of ablation studies to validate the efficacy of our proposed data value on both levels. Initially, to verify the effectiveness of task value estimation, we standardize the weight of all tasks to 1 and then conduct data selection based on instance influence only. Subsequently, to verify the effectiveness of instance value estimation, we calculate task weights based on task difficulty, but select instances within task instructions randomly. We present our results in~\autoref{tab:ablation-1}. 

We discover that data selection based on task value alone or instance value alone can both boost the performance on all three benchmarks. And selecting data based on both instance influence and task difficulty achieve the best results than all other baseline methods on all of the benchmarks, which proves the effectiveness of both values.

\begin{table*}[t]
  \caption{The ablation of the effectiveness of different data values.
  }
  \label{tab:ablation-1}
  \centering
  \scalebox{0.85}{
  \begin{tabular}{>{\centering\arraybackslash}p{.20\linewidth}|>{\centering\arraybackslash}p{.14\linewidth}>{\centering\arraybackslash}p{.24\linewidth}*{2}{>{\centering\arraybackslash}p{.17\linewidth}}}
    \toprule
    Benchmarks     & Ours~(Both) & $\neg$ Instance-level &  $\neg$ Task-level &  Neither\\
    \midrule

    SQA-I   &  \textbf{70.6} &  \underline{69.8}  &  68.2    &  68.4  \\
    MMB    &  \textbf{65.0}  &  \underline{63.7} &  62.9    &  62.5  \\
    SEED-I  &  \textbf{63.2}  &  62.7  &  \underline{62.9}   &  62.2  \\
    
  \bottomrule
  \end{tabular}}
\end{table*}

\begin{figure}[tb]
\centering

\begin{subfigure}[t]{.45\textwidth}
        \centering
        \includegraphics[width=.9\textwidth]{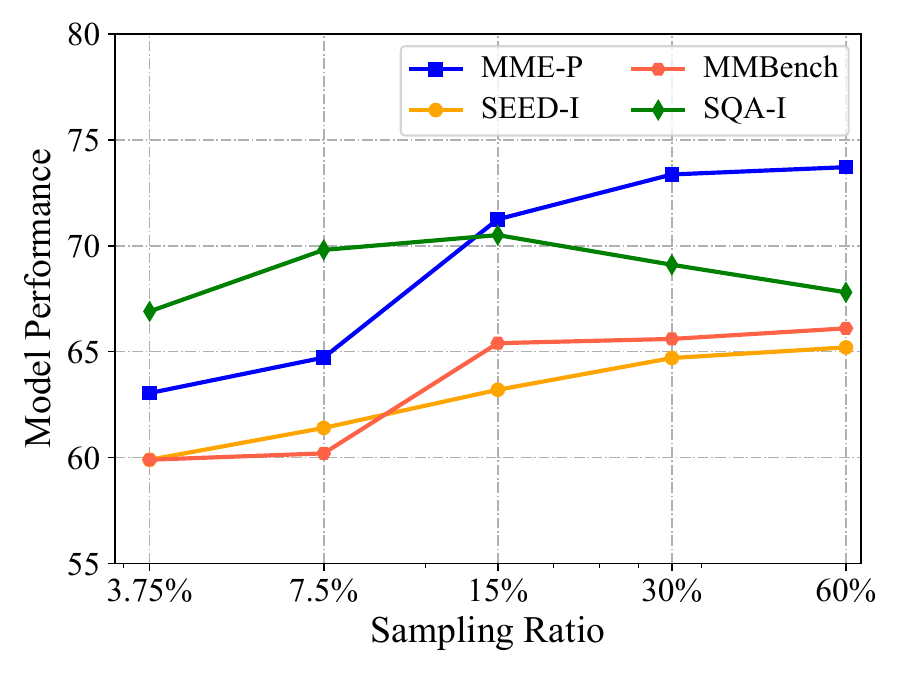}
        \caption{The ablation of different sampling ratio.}
        \label{figure-ablation-2}
\end{subfigure}
\hfill
\begin{subfigure}[t]{.45\textwidth}
    \centering
    \includegraphics[width=.9\textwidth]{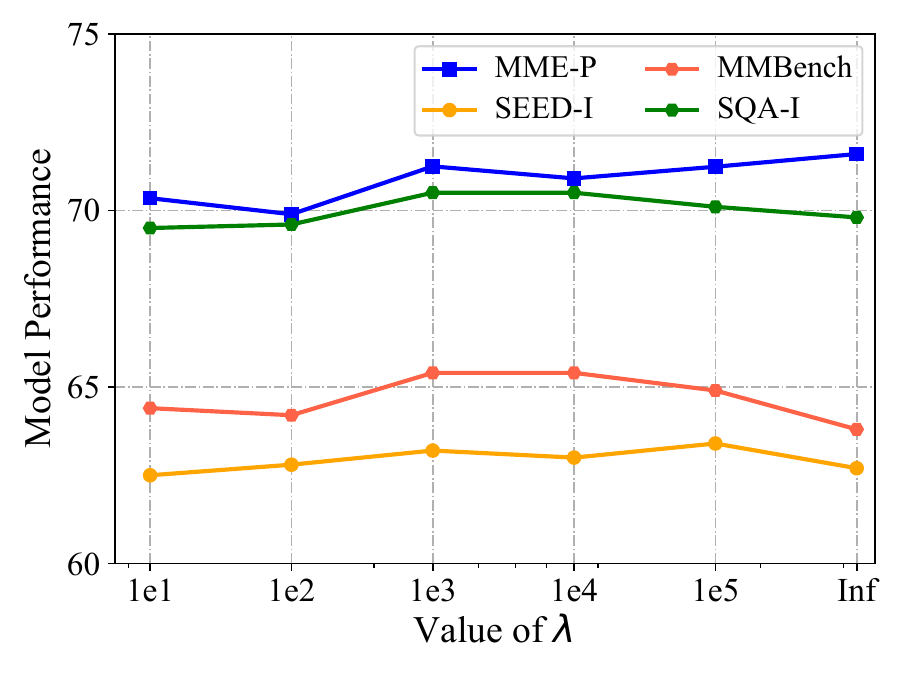}
    \caption{The ablation of selecting data with different hyperparameter $\lambda$.}
    \label{figure-ablation-3}
\end{subfigure}
\caption{The results of ablation study about the data size and hyperparameter $\lambda$.}
\end{figure}

\paragraph{Model Performance with Different Sampling ratio.} To explore the trend of model performance as data size changes, we conduct a series of experiments with different data sampling ratio. In all experiments, we maintain consistency in the data selection approach as well as model training configuration. Our experimental results are presented in~\autoref{figure-ablation-2}. 

As we can observe, the model's performance continuously improves with the increasing amount of data yet, the trend of this enhancement varies across different tasks. The model's performance on MME-P rapidly increases as the data size increases. However, on MMBench and SQA-I, the model's performance increases at first and then stabilizes. A possible reason for this is that MME-P tends to evaluate the model's ability on visual recognition while the other two benchmarks focus on the model's general reasoning capability. Furthermore, We find that the model can maintain a certain level of performance under the minimal data size, indicating that models can acquire basic capability for downstream tasks even with a minimal amount of data.

\begin{table*}[t]
  \caption{The ablation of different warm-up data size. \# Avg indicates the average performance on the benchmarks. We normalize the scores on MME-P and MME-C for computing average performance.
  }
  \label{tab:ablation-5}
  \centering
  \scalebox{0.85}{
  \begin{tabular}{>{\centering\arraybackslash}p{.08\linewidth}|*{9}{>{\centering\arraybackslash}p{.09\linewidth}}}
    \toprule
    \# Samp   &  MME-P  &  MME-C  & SEED-I & MMB 
  &  SQA & SQA-I  &  POPE  & \# Avg\\
  \midrule
    2\% &   1424.9      &   284.6  &  63.1
   &  \underline{65.0}   &  72.1  &  \underline{70.4}  &  \underline{85.4}   & 59.6  \\
      4\%  &   \textbf{1441.9}  &   \underline{321.4}  &  \textbf{63.4}  &  64.6  &  71.3  &  69.3  &  85.3  &  59.6\\
      8\%  &  1433.0  &  \textbf{322.1} &  \underline{63.2} &  \textbf{65.1}  &  \underline{72.2}  &  \textbf{70.6}  &  \textbf{85.6}   &  59.9\\
      16\%  &  \underline{1434.3}  &  281.8 &  \textbf{63.4} &  64.8  &  \textbf{72.3}  &  70.1  &  84.4   &  59.5  
       \\
       32\%  &  1431.3  &  317.1  & 63.1 &  64.7  &  72.0  &  70.1  &  85.2 &  59.7  
       \\
    
  \bottomrule
  \end{tabular}}
\end{table*}

\paragraph{Influence of Different Warm-up Data Size.} We design a series of experiments to investigate the influence of different warm-up data size on the performance of TIVE. We simply change the sampling ratio for warm-up data and maintain consistency in other parts of TIVE selection. The results are presented in \autoref{tab:ablation-5}.

As we can observe, as the sampling ratio increases, the model performance initially exhibits a slight increase trend. Then, it begins to oscillate when the sampling ratio reaches 8\%. Even so, the performance differences between various sampling ratios are quite minimal. This implies that a reference model trained with a minimal amount of warm-up data is already effective for TIVE, making the selection process more efficient.

\paragraph{Influence of Different Hyperparameter $\lambda$.} To achieve a balanced choice between data effectiveness and data diversity, we introduce a hyperparameter $\lambda$ to control the temperature of weight distribution. We study the influence of different $\lambda$ on the quality of final selected data. We set $\lambda$ to different values and evaluate the model's performance on downstream benchmarks. 

The evaluation results on MME-P, MMBench and SQA-I are shown in \autoref{figure-ablation-3}. We can observe a consistent slight increase in the model's performance on MME-P benchmark as $\lambda$ increases, indicating that the MME-P benchmark is highly sensitive to instruction diversity, which is consistent with previous conclusions. On the other hand, the performance on SQA-I and MMBench initially increases with the escalation of $\lambda$, then shows a decline once the $\lambda$ reaches $1e3$. The results demonstrate that our approach with $\lambda = 1e3$ is an optimal data selection strategy that balances data effectiveness and data diversity for the model's consistent optimal performance across all downstream tasks.

\section{Related Work}

\paragraph{Visual Instruction Tuning.} Visual instruction tuning is a crucial part of the construction of LVLMs, which aims to enhance the model's ability on instruction following. The collection of visual instructions is essential for visual instruction tuning. Early studies often employ LLMs to synthesize visual instructions. LVLMs trained on these instructions demonstrate promising capabilities in visual conversation and instruction following, but fail to achieve satisfactory performance on academic benchmark~\citep{goyal2017making,schwenk2022okvqa,marino2019ok}. Subsequent studies~\citep{liu2023improved,luo2024cheap,Dai2023InstructBLIPTG} have usually mixed the synthesized visual instructions and instructions from existing academic datasets together as the final instruction data. LVLMs trained on these mixtures of instructions demonstrate exceptional performance in both understanding and generation scenarios. Despite the success, these efforts solely combine all instructions in a simple way, neglecting the potential redundancy within the instructions from different tasks. We investigate the redundancy in existing visual instruction datasets and propose a measurement for data value based on instance influence and task difficulty to reduce redundancy.

\paragraph{Data Selection for Instruction Tuning.}
With the advancement of LLMs, the significance of data selection has become increasingly prominent due to the high training costs. 
As for instruction tuning, LIMA~\citep{zhou2024lima} is the first to demonstrate that instruction tuning can be accomplished with only a small amount of data. \citet{chen2023maybe} further explores the potential of low data usage in task-specific models. Subsequent efforts focus on estimating the importance of an instruction sample. The importance can be estimated based on certain prior characteristics~(\eg length, complexity, diversity)~\citep{liu2023makes,cao2023instruction}, with the assistance of language models~\citep{jain2023llm,liu2023makes,li2023quantity}, by human efforts~\citep{zhuo2024astraios,muennighoff2023octopack}, or using the gradient-based influence estimation on the validation set of the target benchmark~\citep{xia2024less}. Compared to the data selection approach for language instruction tuning, our approach doesn't only rely on prior characteristics of texts, but considers the importance of visual instructions from a holistic perspective of both image and text. Compared to LESS~\citep{xia2024less}, our approach doesn't require data from downstream benchmark, thereby achieving better generalization ability.


\paragraph{Data Selection for Visual Instruction Tuning.}
Fewer studies have been focusing on data-efficient visual instruction tuning. To the best of our knowledge, there are only two studies currently conducted in this area. Among these studies, InstructionGPT-4~\citep{wei2023instructiongpt} selects high-quality instructions based on several metrics designed in their studies and SELF-FILTER~\citep{chen2024your} proposes selecting instruction data with higher diversity and difficulty by training a score-net. Compared to these studies, We are the first to study data selection for a highly complex mixture of visual task instructions, which provides much better results than the candidate datasets from these studies. To handle such complex visual instructions, we propose a gradient-based approach to estimate data value for efficient and effective task learning. With our approach, we accomplish better results compared to previous studies on data selection for visual instruction tuning with our selected data.
\section{Conclusion}

In this work, we focus on the redundancy issue within a mixture of visual instruction datasets that have been widely used for fine-tuning LVLMs. Through our empirical studies, we find that a significant redundancy exists in the mixed visual instruction datasets, with varying redundancy degrees across different task instructions. To eliminate redundancy, we design a novel method namely TIVE, which first estimates data value based on instance influence and task difficulty, then determines the instruction task proportion and selects representative instances to compose a smaller visual instruction subset for training.
Experimental results indicate that, with the help of our data selection method, using only about 15\% data can achieve comparable performance as the full-data fine-tuned model across eight benchmarks, even surpassing it on some of the benchmarks.


\newpage
\bibliography{iclr2025_conference}
\bibliographystyle{iclr2025_conference}

\newpage
\appendix
\section{Training Dataset}\label{apdx-data}
\subsection{Training Dataset for empirical analysis}\label{apdx-data-empical}
The visual instruction dataset used for our empirical analysis is a subset of the original LLaVA-1.5 instructions. We select one dataset for each type of task. The details of these selected datasets are as followed:

\begin{itemize}
    \item \textit{Open-Ended Visual Question Answering~(OE-VQA):} it requires a model to generate natural language answers without predefined options. We select VQAv2~\citep{goyal2017making} since it's one of the most commonly-used OE-VQA dataset.
    
    \item \textit{Multi-Choice Visual Question Answering~(MC-VQA):} it also requires a model to answer visual questions, but only selects the answer from the provided candidate choices. We select the A-OKVQA dataset~\citep{schwenk2022okvqa}.
    \item \textit{Referring Expression Comprehension~(REC):} it requires a model to generate the regional description of the given object or select the correct object based on the given description. We select RefCOCO dataset~\citep{mao2016generation,kazemzadeh2014referitgame}.
    \item \textit{Visual Conversation~(VC):} it requires a model to generate long conversations based on visual content. We select the VC data from instructions of LLaVA-1.0~\citep{Liu2023VisualIT}.
    \item \textit{Image Caption~(IC):} it requires a model to provide an description of the given image. We select CC3M dataset~\citep{sharma2018conceptual} as it is already used for cross-modal alignment training of LLaVA-1.5~\citep{liu2023improved}.
    \item \textit{Textual Conversation~(TC):} it requires the model to generate conversation in a text-only setting. 
    We select ShareGPT~\citep{zheng2023judging}, as it has been widely used in training LLMs.

\end{itemize}

The statistics of our base dataset are presented in \autoref{tab:empirical-statistic}. 

\begin{table*}[htb]
  \caption{Statistics of base training data for empirical studies.
  }
  \vspace{0.05cm}
  \label{tab:empirical-statistic}
  \centering
  \small
  \scalebox{0.9}{
  \begin{tabular}{*{7}{>{\centering\arraybackslash}p{.12\linewidth}}}
    \toprule
    Task     & MC-VQA & OE-VQA & REC & VC & Caption & TC\\
    \midrule
    Numbers. & 60K              & 80K            & 120K                               & 40K                 & 100K    & 40K\\
  \bottomrule
  \end{tabular}}
\end{table*}

\subsection{Training Dataset for main experiments}
We conduct experiments on three datasets. In the experiment of evaluating TIVE against other baseline methods, we adopt the LLaVA-1.5 instruction datasets. In the experiment of evaluating the transferability of TIVE across different datasets, we additionally use the Mini-Gemini and SVIT-Mix instruction datasets. LLaVA-1.5 and SVIT-Mix contains over 600K instructions and Mini-Gemini contains over 1.4M instructions. All these datasets encompass at least nine sub-task datasets. We exclude the caption data from the selection process since it's already been trained during the LLaVA's pre-training stage.

\section{Evaluation Benchmarks}\label{apdx-evaluation}

To comprehensively evaluate the efficacy of our approach, we evaluate TIVE across various benchmarks. The details of these benchmarks are as followed:

\begin{itemize}
    \item \textit{MME:}~\citep{fu2023mme} it evaluates LVLM's reasoning ability from the two dimensions of perception and cognition. Each instance in MME includes an image and two binary questions. We evaluate TIVE on both splits.
    \item \textit{MMBench:}~\citep{liu2023mmbench} it is a systematically-constructed dataset for evaluating the capacity of LVLMs. It encompasses an evaluation of 20 fine-grained capabilities of LVLMs. The evaluation is performed through its official website. We evaluate TIVE on both english split and chinese split to test its multilingual capability.
    \item \textit{SEED-Bench:}~\citep{li2023seed} it develops a comprehensive set of multimodal evaluation tasks across twelve dimensions with the assistence of GPT-4. SEED-Bench encompasses assessments of both image and video understanding capabilities. In our experiments, we only utilize the image benchmark of SEED-Bench.
    \item \textit{ScienceQA:}~\citep{lu2022learn} it is a benchmark constructed around various science topics, encompassing both pure text-based questions and image-related text questions. In our experiment, we assess ScienceQA under both multi-modal and uni-modal setting.
    \item
    \textit{POPE:}~\citep{li2023evaluating}
    it designs a polling-based query approach for the evaluation of object hallucination. It contains 3000 binary questions and support four evaluation metrics. In our experiment, we report the results of accuracy.
    
\end{itemize}

For simplicity, we only adopt MME-Perception, MMBench, and ScienceQA-Image during our empirical analysis.

\section{Baseline Models}\label{apdx-baseline}

We compare TIVE with other baseline models in \autoref{fig:intro} and \autoref{tab:main-table}. These models include: BLIP-2~\citep{li2023blip}, InstructBLIP-7B~\citep{Dai2023InstructBLIPTG}, Shikra~\citep{chen2023shikra}, IDEFICS-80B~\citep{laurenccon2024obelics}, Qwen-VL~\citep{bai2023qwen}, Qwen-VL-Chapt~\citep{bai2023qwen}, InstructionGPT-4~\citep{wei2023instructiongpt}, SELF-FILTER~\citep{chen2024your}, Yi-VL-34B~\citep{young2024yi}, LLaVA-Next-8B~\citep{liu2024llavanext}, Libra~\citep{DBLP:conf/icml/0008YSX24}, and DeepSeek-VL~\citep{lu2024deepseek}.

\section{Implementation Details}\label{apdx-implement}
We utilize Bunny~\citep{he2024bunny} to construct LVLM with different LLM backbone.
We follow the training settings of LLaVA-1.5 across all
experiments. During fine-tuning, the learning rate is set to 2e-5 and the batch
size is set to 16. All models are trained for one epochs. The training settings
for reference models are the same as the previous settings. For all experiments, we sample 8\% of the total instructions and train the reference model on the sampled data for one epoch. We provide a detailed description of our approach in Algorithm \autoref{alg1} and Algorithm \autoref{alg2}.

\begin{algorithm}[t!]
\caption{Estimating Task Difficulty and Instance Influence.}
\label{alg1}
\begin{algorithmic}[1]
{
\REQUIRE \vspace{0.5mm}\texttt{Instruction dataset $\mathcal{D} = \left\{ D_1,...,D_n \right\}\ $}; 
\\
\STATE \vspace{1.mm} \texttt{Training a reference model $M_{\theta}$}; 
\vspace{0.5mm}
\FOR{$D_i \in \mathcal{D}$} 
\STATE \texttt{Initialize task difficulty  } $v_i^t \leftarrow 0$ ; 
\FOR{$s_j \in D_i$} 
\STATE \texttt{Initialize instance  influence  } $v_s^j \leftarrow 0$ ;
\STATE \vspace{0.5mm} $v_i^t 
\leftarrow v_i^t + \nabla l(s_j, \theta) \cdot \nabla l(s_j, \theta)$ ;
\textcolor{teal}{\hfill// Self-influence.}
\FOR{$s_k \in D_i / s_j$}
\STATE \vspace{0.5mm} $v_s^j \leftarrow v_s^j + \nabla l(s_j, \theta) \cdot \nabla l(s_k, \theta) \, / \, |\nabla l(s_j, \theta)| |\nabla l(s_k, \theta)|$ \textcolor{teal}{\hfill//  Influence on other instances.}
\ENDFOR
\STATE  \texttt{Final instance influence } $v_s^j \leftarrow v_s^j \, / \, |D_i| $ ;
\ENDFOR
\STATE  \texttt{Final task difficulty} $v_i^{t} \leftarrow v_i^{t} \, / \, |D_i|$ ;
\ENDFOR
\RETURN $v^t, v^i$
} 
\end{algorithmic}
\end{algorithm}

\begin{algorithm}[t!]
\caption{Data Selection Based on Data Value.}
\label{alg2}
\begin{algorithmic}[1]
{
\REQUIRE \vspace{0.5mm}\texttt{Instruction dataset $\mathcal{D} = \left\{ D_1,...,D_n \right\}\ $}; 
\REQUIRE \vspace{0.5mm}\texttt{Data Value $v^t,  v^i$, pruning ratio $\delta$ }; 
\\
\STATE \vspace{0.5mm} \texttt{Initialize target dataset $\mathcal{D_T} \leftarrow \{ \}$}; 

\vspace{0.5mm}
\FOR{$D_i \in \mathcal{D}$} 
\STATE \texttt{Determine data proportion $|D_i^{'}| = |\mathcal{D}| v^t_i \, / \, \sum_{j=1}^{n}v^t_j$} ;
\STATE \texttt{Map weight distribution $w_i = $ softmax$(v^i\, / \, \lambda) $} ; 
\STATE $D_i^{'} = \texttt{Sample}_{w_i}(D_i, |D_i^{'}|)$ ;  \textcolor{teal}{\hfill//  Sample $D_i^{'}$ based on weights $w_i$.}
\STATE \texttt{Merge into the target data $\mathcal{D_T} \leftarrow \mathcal{D_T} + D_i^{'}$} ;
\ENDFOR

\RETURN $\mathcal{D_T}$} 
\end{algorithmic}
\end{algorithm}

\section{Additional Experiment Details}

\subsection{Calculated Task Proportion}
We present the task proportion calculated via the task-level data value for LLaVA-1.5 instructions in \autoref{tab:apdx1}.

We find that tasks which require precise answers to visually related questions have a relatively higher proportions in the final selected subset. The potential reason is that these tasks often require models to possess a higher level of visual reasoning ability, which contributes more to the enhancement of the model's performance on downstream tasks. Furthermore, task-level values computed by TIVE are consistent with the findings presented in \autoref{empirical}, proving the effectiveness of our method.

\begin{table*}[t]
  \caption{Statistics of calculated task proportion.
  }
  \label{tab:apdx1}
  \centering
  \small
  \scalebox{0.9}{
  \begin{tabular}{>{\centering\arraybackslash}p{.1\linewidth}|>{\centering\arraybackslash}p{.07\linewidth}>{\centering\arraybackslash}p{.06\linewidth}>{\centering\arraybackslash}p{.09\linewidth}
  >{\centering\arraybackslash}p{.11\linewidth}>{\centering\arraybackslash}p{.04\linewidth}
  *{2}{>{\centering\arraybackslash}p{.09\linewidth}}>{\centering\arraybackslash}p{.11\linewidth}
  }
    \toprule
    Task     & VQAv2  & GQA   &  OCRVQA  & A-OKVQA  & VG & RefCOCO    &  ShareGPT &  LLaVA-1.0 \\
    \midrule
    Proportion & 20.0\%             & 12.7\%   &  12.8\%  &  35.3\%  &  4.1\%  &  7.3\%  &  2.2\%         & 5.5\%  \\
  \bottomrule
  \end{tabular}}
\end{table*}

\subsection{Scaling Instruction Numbers across all models}
\begin{wrapfigure}{R}{0.45\textwidth}
     \centering
\includegraphics[width=0.45\textwidth]{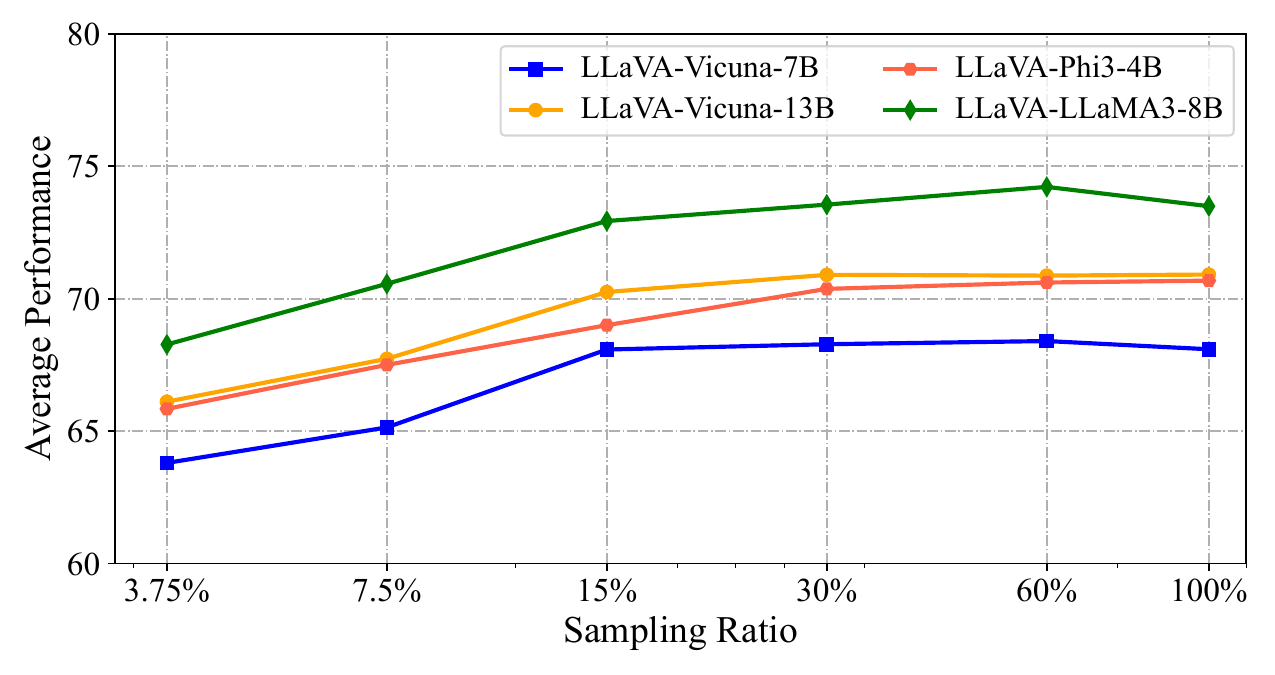}
\caption{The experiment of scaling instruction number on different LVLMs.}
\label{fig:abdx2}
\end{wrapfigure}

To further explore the trend of model performance as data size changes, we conduct the experiments of scaling selected instructions on more models. The data selection approach is still consistent with the previous experiments. The results are presented in \autoref{fig:abdx2}.

We find that the performance of different models follow a similar trend with the increase in instruction number. When the sampling rate is low~(less than 15\%), the performance of all models significantly improves with the increase in instruction number. However, when the sampling rate reaches 15\%, the model's performance gradually stabilizes, scaling instruction number will have minimal effect to the model's performance. Meanwhile, when the sampling rate exceeds 60\%, increasing the number of instructions can even have a negative impact on some models. These experimental results indicate that visual instruction redundancy is clearly present in different models and can potentially have a significant side effect.

\end{document}